\documentclass[runningheads]{llncs}

\usepackage{eccv}

\usepackage{eccvabbrv}

\usepackage{graphicx}
\usepackage{booktabs}

\usepackage{hyperref}

\usepackage{orcidlink}

\newcommand{\bvae}{$\beta$-VAE}
\newcommand{\avae}{Affinity-VAE}
\newcommand{\cnn}{CNN-aVAE}
\newcommand{\splat}{GMD-aVAE}

\makeatletter
\@ifpackagewith{eccv}{review}
{
    \newcommand*{\ack}{}
    \newcommand*{\github}{https://github.com/...}
}
{
    \newcommand*{\ack}{
        \section*{Acknowledgements}This work was supported by the Medical Research Council grant MR/V000403/1 and Ada Lovelace Centre. This work was supported by Wave 1 of The UKRI Strategic Priorities Fund under the EPSRC Grant EP/W006022/1, particularly the “AI for Science” theme within that grant \& The Alan Turing Institute. High performance compute was provided by Baskerville under EPSRC Grant EP/T022221/1. The authors would like to thank James Parkhurst and Joel Greer for data simulation support and Christopher Soelistyo and Alister Burt for technical discussions.
    }
    \newcommand*{\github}{https://github.com/alan-turing-institute/affinity-vae}
}

\makeatother

\begin{document}

\title{{\avae}: incorporating prior knowledge in
representation learning from scientific images} 

\titlerunning{Affinity-VAE}

\author{Marjan Famili{*}\inst{1}\orcidlink{0009-0003-0426-3721}  \and
Jola Mirecka{*}\inst{2}\orcidlink{0000-0001-9361-1713} \and Camila Rangel Smith{*}\inst{1}\orcidlink{0000-0002-0227-836X} \and Anna Kostanska\inst{3}\orcidlink{0000-0001-6377-5477} \and Nikolai Juraschko\inst{1,3,4}\orcidlink{0000-0001-6748-1716} \and Beatriz Costa-Gomes\inst{1}\orcidlink{0000-0002-1073-8442} \and Colin M. Palmer\inst{2}\orcidlink{0000-0002-4883-1546} \and Jeyan Thiyagalingam\inst{2}\orcidlink{0000-0002-2167-1343} \and Tom Burnley\inst{2}\orcidlink{0000-0001-5307-348X} \and Mark Basham\inst{4}\orcidlink{0000-0002-8438-1415} \and Alan R. Lowe\inst{1}\orcidlink{0000-0002-0558-3597}}

\authorrunning{M.~Famili et al.}

\institute{Alan Turing Institute \email{\{mfamili,crangelsmith,bcostagomes,alowe\}@turing.ac.uk} \and
Science and Technology Facilities Council \email{\{jola.mirecka,colin.palmer,t.jeyan,tom.burnley\}@stfc.ac.uk} \and University of Oxford \and Rosalind Franklin Institute \\\email{\{nikolai.juraschko,mark.basham\}@rfi.ac.uk}}

\maketitle

\begingroup
\renewcommand\thefootnote{\textasteriskcentered}
\footnotetext{Equal contribution.}
\endgroup

\begin{abstract}
Learning compact and interpretable representations of data is a critical challenge in scientific image analysis. Here, we introduce Affinity-VAE, a generative model that enables us to impose our scientific intuition about the similarity of instances in the dataset on the learned representation during training. We demonstrate the utility of the approach in the scientific domain of cryoelectron tomography (Cryo-ET) where a significant current challenge is to identify similar molecules within a noisy and low contrast tomographic image volume. This task is distinct from classification in that, at inference time, it is unknown whether an instance is part of the training set or not.  We trained {\avae} using prior knowledge of protein structure to inform the latent space. Our model is able to create rotationally-invariant, morphologically homogeneous clusters in the latent representation, with improved cluster separation compared to other approaches. It achieves competitive performance on protein classification with the added benefit of disentangling object pose, structural similarity and an interpretable latent representation. In the context of Cryo-ET data, {\avae} captures the orientation of identified proteins in 3D which can be used as a prior for subsequent scientific experiments. Extracting physical principles from a trained network is of significant importance in scientific imaging where a ground truth training set is not always feasible.
\end{abstract}

\section{Introduction}

Lying at the core of machine learning research, representation learning is one of the most paramount challenges in our data-driven world. In particular, scientific or image data is often overly redundant and can be meaningfully described using a lower-dimensional representation which captures the salient features of the dataset. The performance of any downstream task, such as clustering or classification has been largely driven by an appropriately encoded (learned) representation of input data \cite{Kingma2014}. In real-life or scientific scenarios obtaining ground truth annotations is often costly, laborious or otherwise challenging due to a lack of information. The development of new methods for interpretable, factorized representations of data without direct supervision is therefore crucial to future scientific discoveries.

Variational autoencoders \cite{Kingma2014}, such as {\bvae} \cite{Higgins2017} have advanced various visual recognition tasks, by learning compact low-dimensional representations of the distribution of the data. The assumption is that these methods can uncover the key generative factors of the data which encode semantic context. However, learning continuous latent representations in volumetric image data that separate translation, rotation, and object semantics, remains a challenge. Despite significant developments, such representations lack control over learned semantics due to their data-agnostic nature \cite{Bepler2019explicitly,locatello2019,Ziatdinov2021}. By introducing domain-appropriate inductive biases in the model or data, through prior knowledge, it is feasible to learn useful scientific models from a corpus of data \cite{Soelistyo_2024_CVPR}. In domains like visual tasks, leveraging common characteristics such as object affinity could enhance representation learning \cite{neuro,neuro2}.

As an exemplar, we use an open scientific challenge: identification of target molecules in volumetric cryogenic Electron Tomography (cryo-ET) image data \cite{cryoET}. Cryo-ET is an emerging high-resolution imaging technique that has the potential to revolutionize our understanding of molecular and cellular biology. It enables the three-dimensional spatial distribution of the entire proteome of molecules inside whole cells to be resolved, with near-atomic resolution \cite{tomo,proteo,proteomics}.  Cryo-ET tomograms are generated by collecting a tilt series of a frozen specimen in a transmission electron microscope. The aligned 2D projection images are back-projected to create the 3D tomogram \cite{Turk2020}, enabling resolution of the entire visible proteome in 3D \cite{plates,detectors}. Advances in instrumentation have not been complemented by equivalent methodological developments for contextual information extraction from reconstructed volumes \cite{proteo} which includes  particle recognition, classification, and subtomogram averaging to enhance local resolution and signal-to-noise ratios (SNR) \cite{resolution,Castano2019,Wan2016}. However, challenges such as low SNR, molecular crowding, and conformational heterogeneity persist, driving the need for computational strategies for subtomogram target identification.

\begin{figure}
    \centering
    \includegraphics[width=0.9\textwidth]{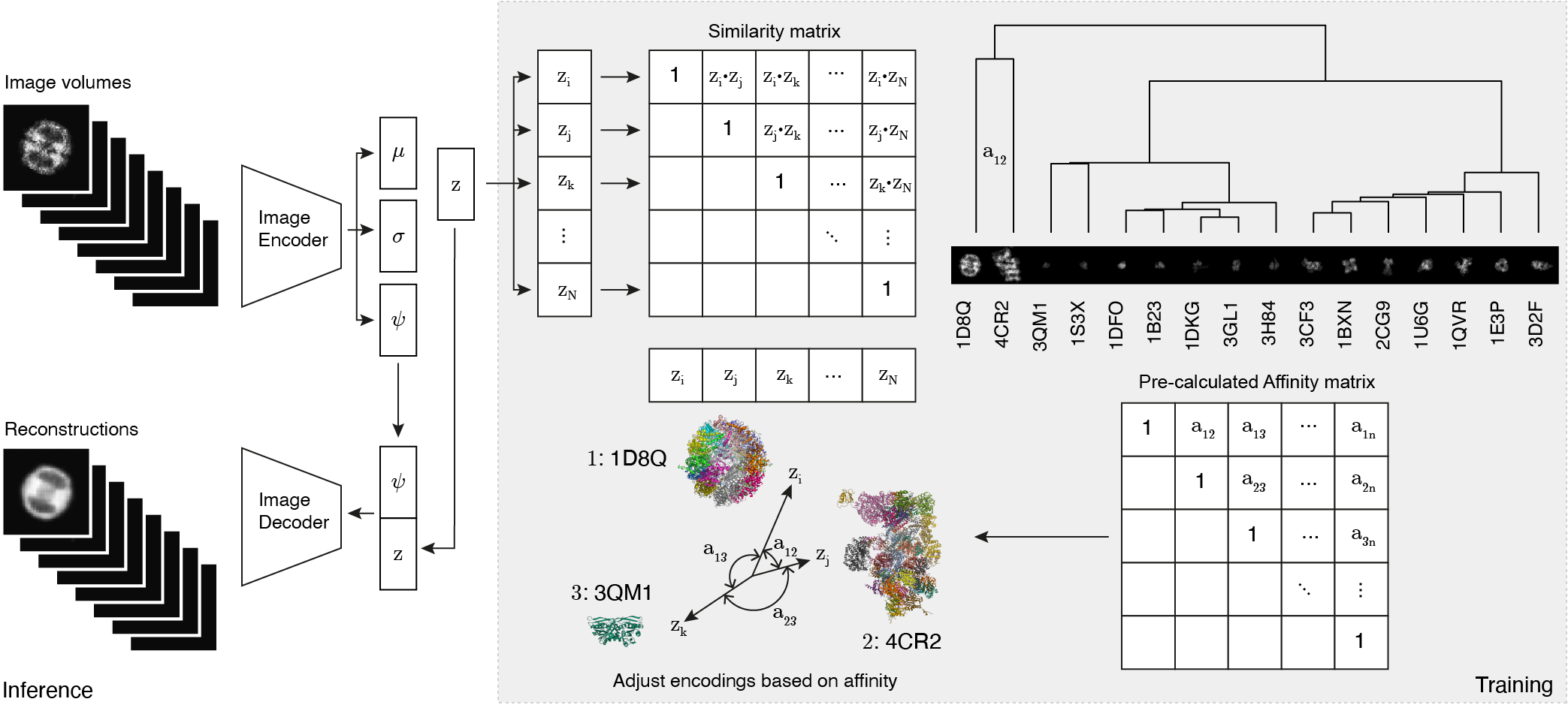}
    
    \caption{\textbf{Architecture of Affinity-VAE}. The mini-batch of training examples (which can be either 2D or 3D images) is encoded by the image encoder as vectors of latent representations of the underlying shape and corresponding vectors representing the intra-class variance arising, for example, from differences in rotational pose ($\psi$). The concatenated latent representations ($\mathbf{z}$) and poses ($\psi$) are used by an image decoder to reconstruct the outputs.  During training, a pre-computed affinity matrix, representing the shape similarity score between the known object classes ($Y$) in the training dataset ($X$) is used to adjust the latent encodings ($z_1, \ldots,~ z_N$) to better represent our prior knowledge of their shapes.  At inference time, Affinity-VAE uses the latent representation to encode and classify unseen objects based on their affinity with the training classes. Four character codes refer to unique PDB accession codes. }
    \label{pipeline}
\end{figure}

 Our proposed model, the Affinity Variational Autoencoder (Affinity-VAE, Fig. \ref{pipeline}), is an enhancement to {\bvae} by incorporating prior knowledge into the regularization of the latent space. This method improves clustering and classification of unseen objects based on their similarity to the classes in the trained model. We achieve this by disentangling features relating to the shape and pose of objects during training, ensuring that the learned latent representation captures only the morphological features of the underlying molecule. Simultaneously, we penalize the loss function in accordance with the cosine similarity of the latent vectors and the pre-calculated affinity score. This ensures that the adjacency of classes in the latent space is in accordance with our affinity score, enabling inference of scientific meaning from the placement of previously-unseen objects in the representation space.

\section{Related Work}

Although substantial research has been performed on the regularization of the latent space in VAEs to incorporate prior knowledge such as class conditioning \cite{Sohn2015cVAE}, here we explore the use of problem specific information to disentangle shape from other sources of intra-class variance such as rotational pose. Convolutional Neural Networks (CNNs) have widely been used in the context of cryo-EM and cryo-ET \cite{deeptomo,Bepler2019,Wagner2019,Palovcak5015,Bepler2019explicitly,Zhong2021,dynamight,Zeng2023,Kimanius_2022_NeurIPS,Chen_2021_GMM} owing to their efficiency in  reducing the dimensionality of 2D/3D images by encoding them into a low-dimensional latent representation. This approach has proven successful in characterising macro-molecules, membranes, and non-cellular features in tomograms \cite{vaetomo}. 

Further exploration into rotationally invariant VAEs such as rVAE/jrVAE \cite{Ziatdinov2021,ziatdinov_kalinin_2021} and rotationally-equivariant spherical CNNs \cite{Esteves2019Spherical} has led to a growing interest in novel methods in this field. Notably, similar to our work, Spatial-VAE \cite{Bepler2019explicitly} explores disentanglement of pose generative factors in an unsupervised manner through specific regularization of pose-related features. However in contrast to our approach, Spatial-VAE is not concerned with imposing knowledge-based latent regularization.

\section{Methods}

Affinity-VAE allows the incorporation of prior knowledge in the form of a one-off pre-calculated affinity score between the set of classes ($Y$) in the dataset ($X$) during training (Fig. \ref{pipeline}). This affinity score is used for the regularization of latent space to reflect the affinity between the classes. The goal of training in Affinity-VAE is to minimize the loss function $L$,


\begin{equation}
    L(\theta,\phi,\beta,\gamma) = - \mathbb{E}_{\mathbf{z} \sim q_\phi (\mathbf{z} \vert \mathbf{x})} \log p_\theta(\mathbf{x} \vert \mathbf{z}) +  \beta D_{KL}\bigl( q_\phi (\mathbf{z} \vert \mathbf{x}) \parallel  p_\theta(\mathbf{z})\bigr) + \gamma S(\mathbf{z})
    \label{loss_func}
\end{equation}

The first term of the equation is the reconstruction loss minimising the difference between the inputs and decoded outputs.  $~D_{KL}$ refers to the Kullback-Leibler divergence between the posterior ($q_\phi (\mathbf{z} \vert \mathbf{x})$) and  the prior distribution $ \mathcal{N}(0, I)$. This is parameterized by the free parameter $\beta$ to give the variational term which regularizes the representation of the encoded latent space ($\mathbf{z}$).  In addition to the reconstruction and $D_{KL}$ terms of the  {\bvae} \cite{Higgins2017} loss function, we introduced a new affinity regularization term $S$, fine-tuned by the hyperparameter $\gamma$, which reflects the L1 norm of the difference between a pre-calculated affinity matrix $\mathbf{A}$ and the cosine similarity of the latent representations,

\begin{equation}
    S(\mathbf{z}) = \Biggl (
        \sum_{i, j}^{N}\ 
        \left|\left|
            \mathbf{A}_{m n}
            -
            \frac{
                z_i \cdot z_j
            }{
                ||z_i|| \cdot ||z_j||
            }
        \right|\right|
    \Biggr ) / N
    \label{eq-shape}
\end{equation}
with $z$ denoting the latent variables, $i, j$ the indices of the vectors in the mini-batch and $N$  the batch size. Whilst the cosine similarity is used as a measure of similarity between two latent vectors, the pre-computed affinity matrix ($\mathbf{A}$) represents their known pairwise similarity, independent of intra-class variability. We define a surjective function ($f:~X \rightarrow Y$) that maps the set of training examples ($i, ~j \in X$) to the set of class labels ($Y$), and can be used to look up the indices of the pre-calculated affinities between the classes (\textit{e.g.} $(m, n) = (f(i), f(j))$). The concept of similarity can be  established through scientific context. For instance, in the case of proteins, we can integrate more fundamental information  obtained from techniques such as sequence alignment, overlap of atomic potentials or Fourier shell correlations (See \S\ref{sec:similarity-metrics}) through the affinity matrix.

\subsection{Pose channel}
The latent space is represented in terms of distributions expressed through learnable fully connected layers for the mean ($\mu$) and the variance ($\sigma$). We introduced a third learnable fully connected layer to represent the $n$D pose of the object ($\psi$ in Fig. \ref{pipeline}), in a similar manner to rotation in an rVAE~\cite{Ziatdinov2021,ziatdinov_kalinin_2021}. 
By setting the class-based affinity matrix to identity (\textit{i.e.} maximum similarity within the objects of the same class regardless of their orientation), we discourage any within-class variation from the latent space. This forces within-class variation (\textit{i.e.} rotation, translation, \textit{etc}.) to be encoded through free pose parameters in order to preserve the integrity of the reconstruction.

\subsection{Hyperparameters}\label{hparam}

For each set of calculations provided here, we have performed extensive benchmarking for the hyperparameters such as the learning rate, $\beta$ and $\gamma$ as well as network parameters such as latent dimensions and the depth and breadth of the network. The final choice of hyperparameters is heavily dependent on the dataset. 
We use a cyclic scheduling scheme for the hyperparameters $\beta$ and $\gamma$ appearing in Eq. \ref{loss_func}, similar to that described in reference  \cite{fu_cyclical} with a smooth transition between 0 and the maximum value of $\beta$ and $\gamma$. We find that this approach allows for better minimization of the reconstruction loss especially when the data is noisy. In the case of noisy data, changing the contribution of Affinity and $D_{KL}$ losses between a minimum and maximum periodically, allows the network to learn from the reconstruction loss. 

\subsection{Decoder models}

In this study, we used two separate decoder networks, a fully convolutional decoder and a gaussian mixture decoder. The advantage of this gaussian mixture decoder is that there is an explicit constraint on values learned in the pose channel to represent rotation in 3D hence addressing the possibility of shape information leaking to the pose channel. 

\subsubsection{Convolutional decoder} The first is a standard convolutional decoder ({\cnn}) taking  the concatenated latent and pose representations as input. 

\subsubsection{Gaussian mixture decoder}
\label{sec:splat-decoder}

The second is a Gaussian mixture decoder ({\splat}), which places the centre of a Gaussian on each of the $M$ elements (typically $M=1024$) described by the latent vector and applies an explicit rotation to the centroids of the Gaussians, before using 3D Gaussian splatting technique \cite{Chen_2021_GMM,Kerbl2023GaussianSplatting} to reconstruct the images. From the latent encodings and the pose estimation, four sets of parameters are estimated: (i) the covariance of each Gaussian ($\Sigma$). (ii) the weight of each Gaussian ($\phi \in \{0, 1\}$). Note that, since the weight is binary, we use a straight-through estimator \cite{binarized_weight,bengio2013estimating} during training. (iii) an angle-axis representation of the pose ($\psi = \langle \theta, \hat{e}_x, \hat{e}_y, \hat{e}_z \rangle$), by setting the pose channel dimensions to 4D, and (iv) the coordinates of the centroids in the output volume ($\mathbf{v} \in \mathbb{R}^3$). The rotation of the object in $\mathbb{R}^3$ space is represented using a quaternion, $\mathbf{q} =  q_0 + q_1 \mathbf{i} + q_2 \mathbf{j}+q_3\mathbf{k}$. In this representation, $\mathbf{i}, \mathbf{j}, \mathbf{k}$ are mutually orthogonal imaginary unitary vectors and $q_0, q_1, q_2, q_3$ are real numbers. Here we define any axis-angle rotation by four elements $(\theta,\hat{e}_x,\hat{e}_y,\hat{e}_z)$ where our rotation axis is the unit vector $(\hat{e}_x, \hat{e}_y, \hat{e}_z)$ and $\theta$ the angle of rotation. We can convert these values to a rotation quaternion as $\mathbf{q} = \langle \cos(\theta/2), \hat{e}_x \sin(\theta/2),~\hat{e}_y \sin(\theta/2),~\hat{e}_z \sin(\theta/2)\rangle$. The corresponding 3D rotation matrix for the rotation quaternion above is given as:
\begin{equation}
\mathbf{R}_q =\begin{bmatrix}
1-2(q_{2}^{2}+q_{3}^{2})& ~2(q_{1}q_{2}-q_{0}q_{3})& ~2(q_{1}q_{3}+q_{2}q_{0})\\
2(q_{1}q_{2}+q_{3}q_{0})& ~1-2(q_{1}^{2}+q_{3}^{2})& ~2(q_{2}q_{3}-q_{1}q_{0})\\
2(q_{1}q_{3}-q_{2}q_{0})& ~2(q_{2}q_{3}+q_{1}q_{0})& ~1-2(q_{1}^{2}+q_{2}^{2})
\end{bmatrix}
\end{equation}

which can be applied to the encoded points, using the transform $\mathbf{v}^\prime = \mathbf{R}_q \mathbf{v}$. Finally, our output image volume is a 3D coordinate system $\mathbf{r} = [-1, 1]^3 \subset \mathbb{R}^3$ with N voxels in each dimension. We can calculate the final density using a Gaussian Mixture Model over the entire output volume:
\begin{equation}
y_j = \sum_{i=1}^{M}\phi_i\mathcal{N} \bigl(\parallel \mathbf{v}^\prime_i - \mathbf{r}_j \parallel, \Sigma_i \bigr) ~\forall j
\end{equation}

One can add convolution layers to the end of this decoder, which can be used to learn a transform to convert the density to an image. 

\subsection{Affinity metrics} \label{sec:similarity-metrics}

Regularization of the latent space with respect to the pre-calculated affinity score ensures the adjacency of the  encoded  objects  within the latent space is in accordance with the provided similarity metric.  To achieve this regularization we require  a pairwise similarity between all classes in training examples. The affinity between two structures can be described in accordance with the intrinsic properties of the data.
The affinity score in our calculations reflects the pairwise shape similarity between classes. We have constructed the shape similarity matrix comparing two methods, SOAP and FSC, as well as hand manipulation to ensure the most meaningful similarity between the classes. The choice of affinity descriptor (shape, or indeed any other metric) should therefore be made with respect to the property of the data intended to achieve the desired data separation.

\subsubsection{Smooth Overlap of Atomic Positions (SOAP)} This descriptor uses a combination of radial and spherical harmonics \cite{Bartok2010,Bartok2013}. The SOAP descriptor places a Gaussian density distribution at the location of each atom. The SOAP kernel is then defined as the overlap of the two local nearest neighbouring densities integrated over all three-dimensional rotations.

\subsubsection{Fourier Shell Correlation (FSC)} This descriptor calculates the correlation between the Fourier coefficients of each image in thin spherical shells \cite{fsc}. We take an average of the FSC across all spatial frequency shells, weighted by the number of Fourier coefficients in each shell according to the method described by Brown \textit{et al.}~\cite{avgfsc}. 

\subsection{Latent embeddings, clustering and classification}

The encoded samples are represented in the latent space through their generative factors, however the dimensionality of that latent space ($z$) is a hyperparameter. We are therefore analysing $n$D latent spaces such that $n$ is high enough to capture the complexity in the original image, but low enough to only capture the true semantics. Therefore, even with the simplest 3D datasets $n > 2$. In order to visually inspect the learned latent manifold we used UMAP \cite{umap} and t-SNE \cite{tsne} dimensionality reduction methods. For each embedding we have chosen the method which best illustrates the underlying features in the latent space.  For the classifications presented in the confusion matrices, we used a multi-layer perceptron neural network classifier. Other algorithms for classification such as  K-Nearest Neighbors (KNN) algorithm \cite{knn}  are also available in the Affinity-VAE repository for validation purposes. 

\subsection{Implementation and training}
Affinity-VAE is implemented in PyTorch and was trained using NVIDIA A100 80Gb GPUs. Source code is available at: \url{\github}

\section{Results}\label{sec2}

In systematic steps, we increase the complexity of our task to demonstrate the efficacy of Affinity-VAE. We start with a simple dataset of 2D images of alphanumeric characters, where the only intra-class variation is rotation, to demonstrate the concept of Affinity-VAE. 
Next, we benchmarked the performance of Affinity-VAE for Cryo-ET data using the SHREC 2021 challenge dataset \cite{shrec}. Initially, we focus on the ground truth (without noise) where subtomograms contain single protein particles and background, introducing more sources of intra-class variation. Finally, we demonstrate the performance on subtomograms extracted from the full reconstruction of tomograms including noise and microscope optics illustrating the potential for the method on challenging experimental datasets. 

\subsection{Alphanumeric dataset}

\begin{figure}
\centering
\includegraphics[width=\textwidth]{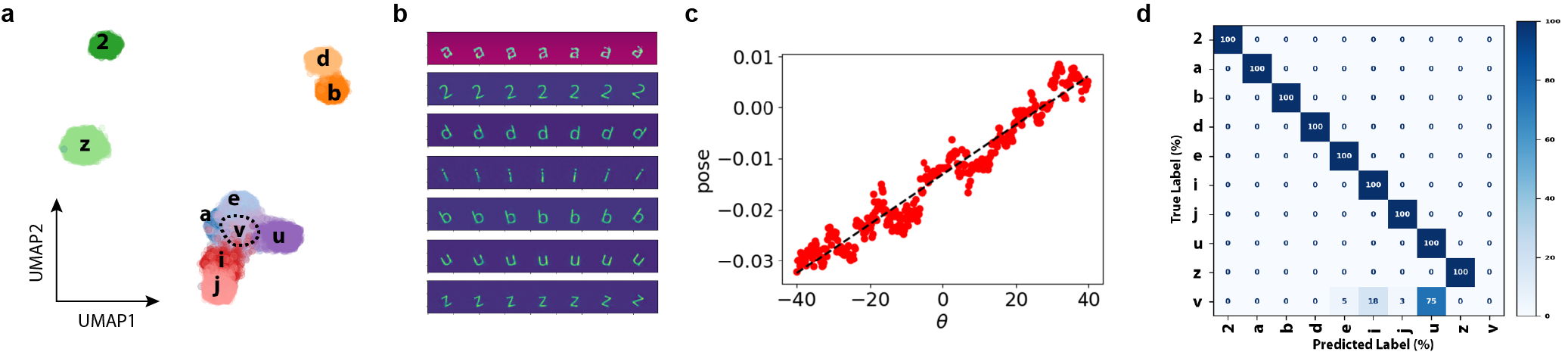}
\caption{The performance of {\cnn} using a simple alphanumeric dataset with rotation as the source of intra-class variation. (a) UMAP embedding of latent vectors of the validation set built from the prediction for 200 samples, with random rotations, of seen (\texttt{a}, \texttt{b}, \texttt{d}, \texttt{e}, \texttt{i}, \texttt{j}, \texttt{z}, \texttt{2} and \texttt{u}) and unseen data (\texttt{v}). (b) Interpolation of the pose channel for an average point in the latent space (pink) and the interpolation of the pose channel conditioned on individual classes. (c) The correlation of the inferred
1D pose with the angle of rotation ($\theta$) of the input image. (d) The confusion matrix for classification of the validation set.}
\label{fig:alphanumeric}
\end{figure}

The alphanumeric dataset is a 2D synthetic dataset, constructed of images of letters and digits ($x \in \{ \texttt{a,e,b,d,i,j,z,2,u,v}\}$). The images are rotated by a random angle of  $\theta$ where $\left\{\theta \in \mathbb{Z}|-45<\theta<45 \right\}$. Since there is no variation in the shape of a given character, the only intra-class variation in this dataset is rotation which allows us to explore the learning of pose. Letter \texttt{v} is unseen by the network during training and reserved for evaluation of the model. Fig. \ref{fig:alphanumeric} shows the ability to disentangle pose from shape, and the classification of the unseen letter \texttt{v} is in accordance with our expectation, mostly classed as \texttt{u}. The characters in this simple dataset are chosen to demonstrate the ability of Affinity-VAE to ignore pose and prioritize shape similarity in organizing the latent space representation.

\subsection{Simulated Cryo-ET tomograms from SHREC 2021 dataset}
\begin{figure}
\centering
\includegraphics[width=0.95\textwidth]{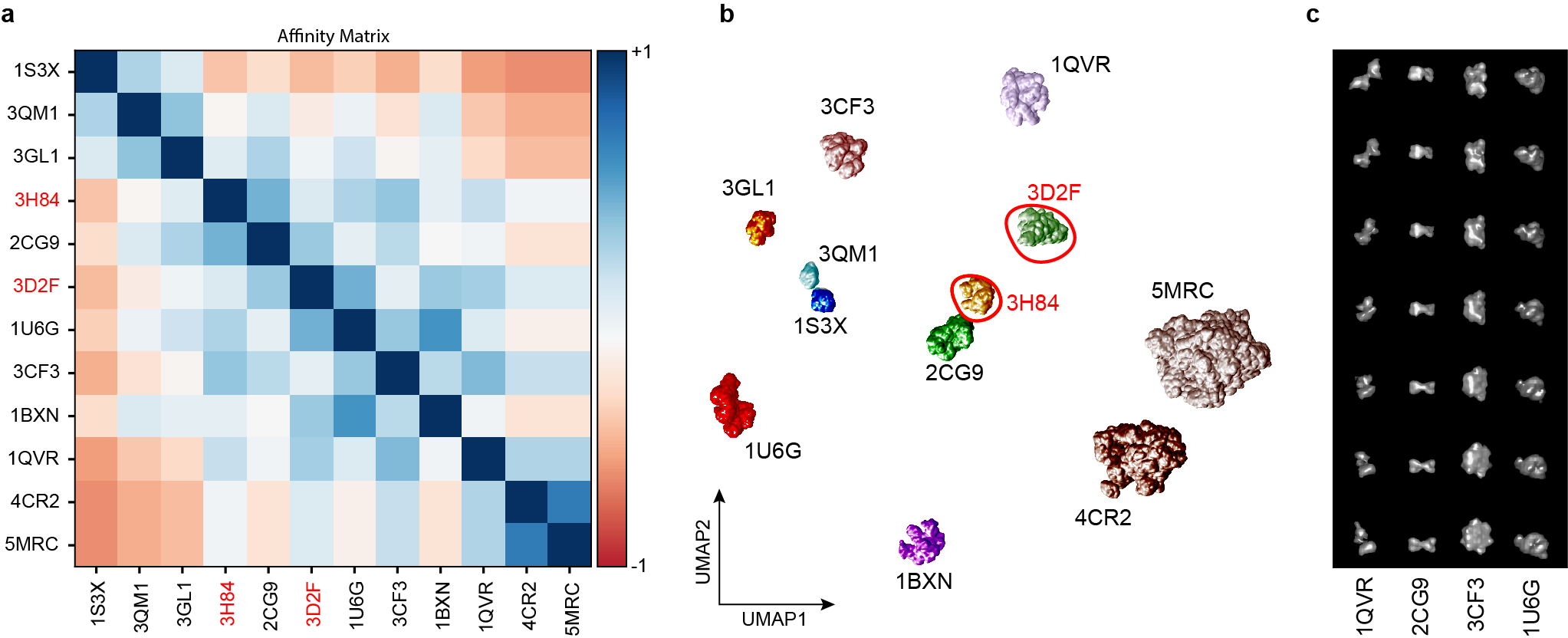}
\caption{Learning a representation of protein shapes. (a) The Affinity matrix calculated using SOAP descriptors. (b) UMAP embedding of latent vectors from {\cnn} where \texttt{3D2F} and \texttt{3H84} are unseen during training and only used for evaluation. The placement of these proteins in the latent space is as expected from their shape similarity to the training classes. (c) Interpolation within the 1D pose channel for four example proteins.}
\label{fig:shrec_single_particle_conv}
\end{figure}

The SHREC 2021 dataset contains 10 tomograms ($512\times512\times512$, with a sampling of 1 nm/voxel), which are split 80:20 for training and evaluation. The tomograms include 12 different proteins ($\sim2\times10^3$ proteins per tomogram) of varied molecular weight and shape \cite{shrec} as well as fiducial markers and membranes. The number of instances per class in the dataset is not identical. We computed an affinity matrix for all proteins using  SOAP descriptors (see \S\ref{sec:similarity-metrics} \& Fig. \ref{fig:shrec_single_particle_conv}) and used this prior for all experiments with the SHREC dataset. We use three variants of the image dataset for experiments: (i) Single proteins only. Here, subtomograms were extracted with ground truth co-ordinates but the crowding from the background was removed. (ii) Subtomograms from the ground truth. Here, each subtomogram represents the molecular crowding observed in real images, having a crop from the larger simulated volume centered on a single protein.  The image represents the electrical charge density of the sample. (iii) Subtomograms from the full simulated reconstruction.  This is similar to (ii) but incorporating the full optics, noise model and reconstruction artifacts of a real Cryo-ET experiment.

Using these datasets, we demonstrate the interpretability of the representation learned by Affinity-VAE, as well as the associated task of protein classification, and compare performance to other competitive methods from the literature. We explored the performance of both our models ({\cnn} and {\splat}) in a systematic study of proteins augmented through rotation, the ground truth and the reconstructed Cryo-ET tomograms including noise and reconstruction artifacts.

\subsubsection{Single proteins only}

We trained the {\cnn} model with proteins in the SHREC dataset where particles are augmented by rotation in 3D through $360$ degrees. Two proteins 3H84 and 3D2F are excluded from training used only for evaluation. Comparing the affinity matrix and the latent embedding indicates that the latent representation conforms closely to the provided affinity matrix. The continuous latent space, regularized by shape similarity, shows that the unseen proteins are placed close to training examples with which they share high similarity, in accordance with our own inductive biases (Fig. \ref{fig:shrec_single_particle_conv}). In addition to visualizing the latent space with dimensionality reduction methods, we can also compare the learned representation to the pre-calculated Affinity matrix, for example comparing Fig. \ref{fig:shrec_single_particle_conv}a \& Fig. \ref{fig:grandmodel_noise}a. Here we observe that the cosine similarity between the latent vectors closely resembles the prior, indicating that the model has learned to map from the image data to a representation that captures our prior knowledge. 

\begin{figure}
\centering
\includegraphics[width=0.95\textwidth]{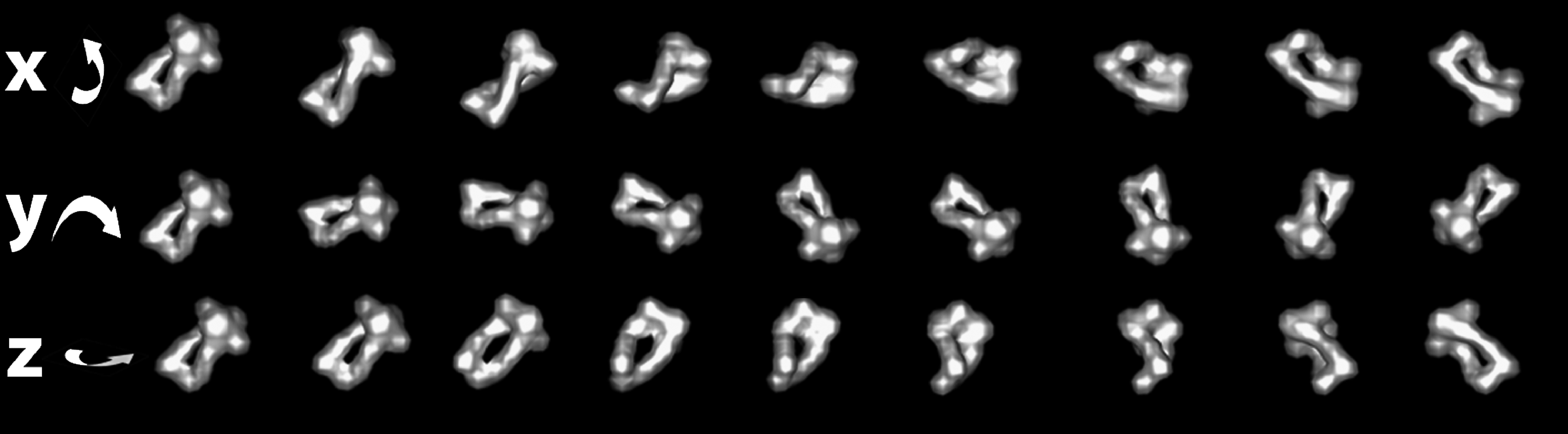}
\caption{Decoded images interpolating within the pose channel for the embedding of \texttt{2CG9} using the Gaussian mixture decoder.}
\label{fig:shrec_single_particle_sgd}
\end{figure}

We also interpolate across pose values for given class latent codes demonstrating consistent class characteristics for each latent representation. In the {\cnn} architecture, we deliberately chose a 1D pose channel to capture the 3D rotations of the objects. Increasing its dimensions can lead to the leakage of information from the latent space, particularly inter-class shape variation, into the pose channel. The decoder uses both channels and since pose space is completely unregularized it is possible that the reconstruction can become independent of the latent content and that the pose channel will not explicitly represent 3D rotation. This issue can easily be resolved by replacing the convolutional decoder with a Gaussian mixture decoder ({\splat}). This ensures that the learned pose always represents rigid-body 3D rotation. Since all variables in this process are differentiable, they can be optimized with standard gradient descent based optimization methods. We also show that interpolating along the learned pose parameters provides smooth transitions between angles of rotation in expected planes. (Fig. \ref{fig:shrec_single_particle_sgd}). 

\begin{figure}
\centering
\includegraphics[width=0.95\textwidth]{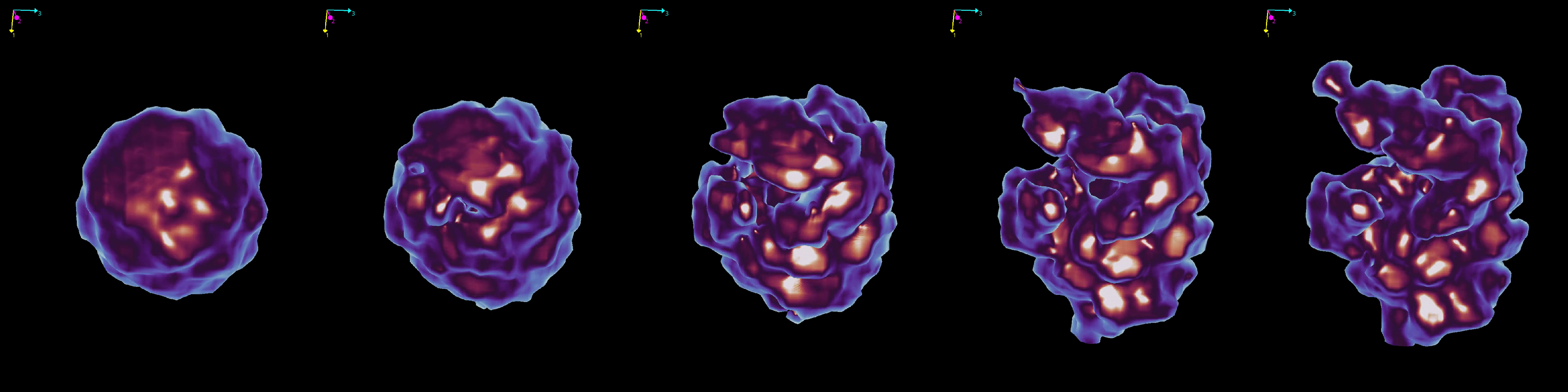}
\caption{Linear interpolation between the latent encodings of \texttt{1D8Q} (left) and \texttt{4CR2} (right) using a Gaussian mixture decoder.}
\label{fig:latent_traversal}
\end{figure}

Further, given two latent encodings for known proteins, we can linearly interpolate in latent space, and decode to image volumes, in order to validate the smoothness and continuity of the representation (Fig. \ref{fig:latent_traversal}). The {\splat} model is able to generate scientifically plausible intermediate structures given the new latent codes allowing for discovery of unseen classes and their appropriate semantic encoding.

\subsubsection{Subtomograms from ground truth}

\begin{figure}
\centering

\includegraphics[width=0.95\textwidth]{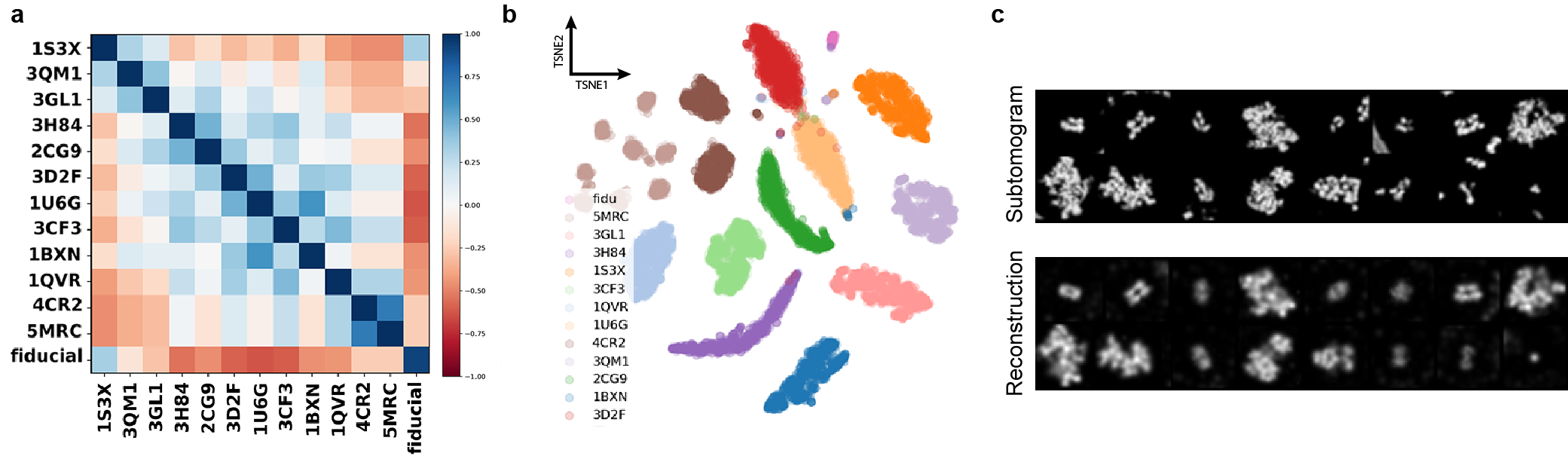}
\caption{Classification of proteins using the ground truth tomograms. This dataset includes a simulated ``fiducial'' class representing non-protein, electron dense fiducial markers used in real experiments for image registration.  (a) Cosine similarity of the latent vectors (b) Latent embeddings. (c) Subtomograms and their reconstructions using the {\splat} network. }
\label{fig:grandmodel_noise}
\end{figure}

To explore the efficiency of the {\splat} in denoising and ignoring class-specific variations, we next look at the ground truth of the SHREC dataset. This includes the subtomograms extracted from the ground truth tomograms where the background of the proteins (simulating molecular crowding in real samples) is an added challenge. In Fig. \ref{fig:grandmodel_noise} we show the organization of the latent space as well as the quality of the reconstructions with the {\splat} model. The latent embedding and the cosine similarity of the latent vectors demonstrate that the model has learned a compact representation of the central protein shapes and has successfully removed the background of the peripheral proteins in the reconstructions. However, although this model is able to efficiently disentangle the shape and pose of molecules in the ground truth dataset, real experimental data has more complex noise characteristics, and the tilt-series acquisition and the microscope's contrast transfer function (CTF) affect the appearance of the molecules.

\subsubsection{Subtomograms from full reconstruction}

\begin{figure}
\centering
\includegraphics[width=0.95\textwidth]{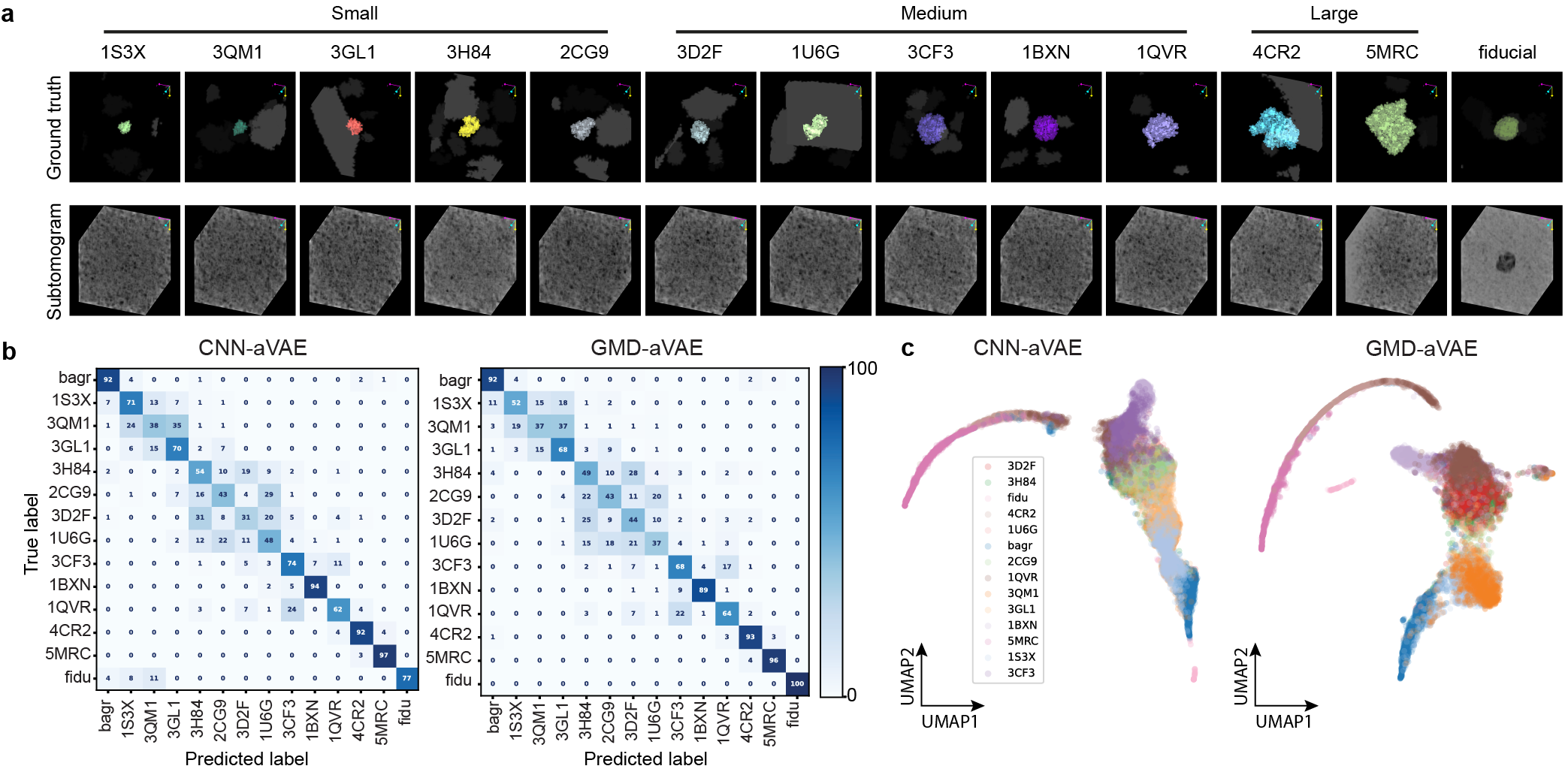}
\caption{Performance of {\avae} when evaluating simulated subtomograms from the fully reconstructed tomograms 8 and 9 of the SHREC2021 dataset. (a) Examples of subtomograms extracted from the full tomogram for each class. Both the ground truth and simulated subtomogram data are shown for each. In the ground truth volumes, density belonging to the non-target class is shown in gray. Simulated subtomograms are minimum intensity projections of the extracted cube. (b) Confusion matrices showing the performance of the decoder models on the test dataset of simulated subtomograms. (c) Latent embeddings. The full dataset also includes a background class (\textit{i.e.} no protein in subtomogram) labelled as ``bagr''.}
\label{fig:shrec_topaz}
\end{figure}

Finally, we illustrate the performance of {\cnn} and {\splat} in learning the pose, affinity scores and classification of proteins using subtomograms extracted from the fully reconstructed tomograms of the SHREC dataset (Fig. \ref{fig:shrec_topaz}a). We first applied the denoising algorithm Topaz \cite{topaz} to the training and evaluation tomograms.  We then performed careful benchmarking to consider the optimal hyperparameter values for $\beta$, $\gamma$ and learning rate of the optimizer as well as the network architecture including pose and latent dimensions. Because the full reconstruction exhibits noise, missing wedge artifacts and the effects of the microscope CTF, we added a learnable convolutional layer after the GMM in the {\splat} model. This convolutional layer effectively simulates the forward transform of the electron density to an observed image in the microscope. After training, both {\cnn} and {\splat} models can disentangle the classes within the latent space and the organization of the classes within the latent space reflects the known affinity scores. Both models achieve an average accuracy of 67\% in the classification task. This is significantly better than a vanilla {\bvae} trained with the same data, which achieves an average accuracy of 50\%. We compared the performance of these models to standard template matching (TM) algorithms, the DeepFinder algorithm \cite{deeptomo} and the best performing models from the SHREC 2021 challenge (Table \ref{table:results}).

\begin{table}
\centering
\caption{Results from SHREC2021 challenge data. The ``Small'' subset represents <200 kDa, \textit{i.e.} \texttt{1S3X, 3QM1, 3GL1, 3H84, 2CG9}, ``Medium'' is 200-600 kDa, \textit{i.e.} \texttt{3D2F, 1U6G, 3CF3, 1BXN, 1QVR} and ``Large'' is >600 kDa, \textit{i.e.} \texttt{4CR2, 5MRC}, as defined in the SHREC challenge documentation. $^\dagger$ This is the best performing model overall for each subset of molecules. The $F_1$-score is calculated as harmonic mean of the precision and recall, and measures predictive performance.}
\begin{tabular}{lccc}
\toprule
 Method  & \multicolumn{3}{c}{$F_1$-score} \\
 & Small & Medium & Large \\
\midrule
 TM & 0.233 & 0.542 & 0.903 \\
 DeepFinder \cite{deeptomo} \quad & 0.563 & 0.882 & 0.985 \\
 Best$^\dagger$ & 0.594 & 0.893 & 0.998 \\
 {\cnn} & 0.531 & 0.617 & 0.936 \\
 {\splat} & 0.517 & 0.609 & 0.945 \\
\bottomrule
\end{tabular}
\label{table:results}
\end{table}

Affinity-VAE achieves competitive performance in the task of classifying proteins in the SHREC2021 challenge dataset. The performance of both models is similar to other state-of-the-art methods, with some notable differences. Inspection of the confusion matrices and latent spaces reveals some interesting features of the learned representation (Fig. \ref{fig:shrec_topaz}). In the SHREC challenge, the dataset was subdivided according to molecular size (small, medium and large, Fig. \ref{fig:shrec_topaz}a). However, because we use the pre-calculated affinity matrix (Fig. \ref{fig:shrec_single_particle_conv}a) to organize the latent representation, we find that {\avae} has grouped proteins based on size \emph{and} structural similarity. Four distinct groups can be observed: (i) \texttt{1S3X}, \texttt{3QM1}, \& \texttt{3GL1}, (ii) \texttt{3H84}, \texttt{2CG9}, \texttt{3D2F} \& \texttt{1U6G}, (iii) \texttt{3CF3}, \texttt{1BXN}, \& \texttt{1QVR} and (iv) \texttt{4CR2} \& \texttt{5MRC} which may explain why the $F_1$ metrics do not capture all the salient properties when comparing models. For example, when comparing instead the smallest group (i), comprising only \texttt{1S3X}, \texttt{3QM1}, \& \texttt{3GL1}, {\splat} outperforms all other models with an $F_1$-score of 0.56 versus 0.47 for DeepFinder, suggesting that {\avae} is particularly performant on very small proteins. The reduced $F_1$-scores (indicating worse performance than SOTA models) for the ``Medium'' proteins originates from two trade-offs: (i) the intentional organization of the latent space by a scientific concept of choice, in this case protein shape similarity, and (ii) capturing the exact pose of the object.  The limited size of the training dataset likely makes the joint optimization very challenging (see Conclusions for more discussion). The results suggest that, given a dataset of sufficient size and diversity, {\avae} is capable of learning a continuous latent representation that can be used for classification, shape inference and pose estimation in challenging Cryo-ET imaging data.

\section{Conclusions}

Here, we present Affinity-VAE, a generative model designed to use prior knowledge to build informed latent representations from scientific image data. We demonstrate the real world utility of the method on a challenging bioimaging problem: identifying protein identity from subtomograms in large volume Cryo-ET imaging data. Our model was trained and validated with the SHREC2021 dataset, and achieves competitive performance on the classification of proteins given subtomogram targets. In addition to pure classification, we performed experiments to determine whether the learned representation enabled protein shape and pose inference. Using held out instances from the dataset, we investigated whether the model could infer characteristics of the underlying molecules. {\avae} was able to correctly place these unseen molecules in the latent space, in proximity to similar structures. Further, by taking advantage of the intermediate representation of the Gaussian mixture decoder we can generate approximate electron densities, yielding an intrinsically interpretable model. This latter feature, interpretable latent representations, is a significant advantage for scientific imaging applications, since we rarely have a training dataset that covers all possible classes. Further, we also show that {\avae} is capable of representing the pose of identified objects in 3D, which will ultimately lead to a dramatic speed up in subtomogram averaging as the time-costly multi-plane alignment can be mitigated, or at least significantly reduced as good initial orientations can provided. Since we are interested in capturing molecular orientation, a future goal is to evaluate the resolution of subtomogram averaged reconstructions of a single class given those pose priors.

Given the promising results, we also note several improvements that could be implemented. First, our method is dependent on the metric used to construct the affinity matrix. Devising appropriate task-specific metrics that capture the scientifically salient features of the dataset, will likely lead to improved results. Chen \textit{et al.} implemented a weakly supervised disentanglement method similar to ours, but explicitly incorporated the variance into a supervised regularization term \cite{chen_2019_weak_disentanglement}. This modification may lead to improved structural inference and classification. Second, in our method we implemented a GMM-based decoder, to enable explicit rotational transforms. However, it is computationally inefficient to evaluate for large 3D volumes as part of a differentiable rendering pipeline. Recently, the authors of CryoDRGN-ET implemented Neural Radiance Fields (NeRFs) for heterogeneous reconstruction \cite{Rangan_2024}, which could be a promising alternative approach. Finally, {\avae} uses convolutional layers to transform the density estimate to an image. Implementing a physics-based forward model of image formation, utilizing an appropriate CTF would likely improve reconstructions, and consequently the utility of the learned representations.

However, the \textit{most} significant factor limiting performance is the paucity of high quality open bioimaging datasets with which to train models such as {\avae}.  The SHREC 2021 dataset is relatively small, with $\sim2\times10^4$ instances total, before a training and validation split. For comparison, modern vision transformer models are typically trained with many millions of examples \cite{dosovitskiy_2020_vit}. In turn, this means that the na{\"i}ve augmentation performed in this and other studies has significant issues due to assumptions made about the microscope optics and CTF. Further, rotation and molecular crowding are not the only sources of intra-class variation in real Cryo-ET data. For example, conformational heterogeneity originates from intrinsic molecular motion, induced conformational changes and interactions with other proteins and cellular structures. Some simulated datasets, combining molecular dynamics and digital twins of the microscope optics are already capturing some of these important details (such as with Parakeet\footnote{\url{https://www.rfi.ac.uk/projects/open-datasets/}} \cite{Parkhurst2021}). Going forward, we anticipate that introducing physical constraints to the model \cite{Esteve-Yague_2023,dynamight} and training on larger, more diverse experimental datasources such as EMPIAR \cite{EMPIAR} and the CZI Cryo-ET data portal\footnote{\url{https://cryoetdataportal.czscience.com/}} will likely lead to improved learned representations and significant performance gains in the future.

\ack

\clearpage

\bibliographystyle{splncs04}
\bibliography{egbib}
\end{document}